\documentclass[conference]{IEEEtran}
\IEEEoverridecommandlockouts

\usepackage{cite}
\usepackage{amsmath, amssymb, amsfonts}
\usepackage{graphicx}

\usepackage{moreverb,url}
\usepackage[colorlinks,bookmarksopen,bookmarksnumbered,citecolor=red,urlcolor=red]{hyperref}
\usepackage{siunitx}
\usepackage{xspace}
\usepackage{tabularx}
\usepackage{xcolor,colortbl}
\usepackage{xparse}
\usepackage{amsthm}
\usepackage{mathtools}
\usepackage{url}
\usepackage{multicol}
\usepackage{soul}
\usepackage{color}

\usepackage{times}
\usepackage{graphicx}
\usepackage{lipsum}
\usepackage{setspace}
\usepackage{epsfig}

\usepackage{amsmath}
\usepackage{bm}
\usepackage{amssymb}
\usepackage{tabularx}
\usepackage{mathtools}
\usepackage{color,soul}
\usepackage{paralist}
\usepackage{tikz}
\usepackage{float}
\usepackage{pdfpages}
\usepackage{makecell}
\usepackage{array}
\usepackage{xcolor}
\usepackage{subcaption}

\usepackage{enumerate}
\usepackage{url}
\usepackage{relsize}
\usepackage{pgfgantt}
\usepackage{amsfonts}
\usepackage{textcomp}
\usepackage{multirow,diagbox}
\usepackage{threeparttable}
\usepackage{booktabs}
\usepackage{bm}

\usepackage{algorithmic}
\usepackage{graphicx}
\usepackage{textcomp}
\usepackage{xcolor}
\def\BibTeX{{\rm B\kern-.05em{\sc i\kern-.025em b}\kern-.08em
    T\kern-.1667em\lower.7ex\hbox{E}\kern-.125emX}}

\newcommand{\etal}{\textit{et al}.~}
\def\figref#1{Fig.~\ref{#1}}
\def\tabref#1{Table~\ref{#1}}

\begin{document}

\newcommand\todos[1]{\textcolor{red}{TODO: #1}} % TODO

\newcommand{\comments}[1]{\textcolor{red}{#1}}

\definecolor{niceorange}{RGB}{230,159, 0}
\definecolor{niceskyblue}{RGB}{86,180, 233}
\definecolor{nicebluishgreen}{RGB}{0,158, 115}
\definecolor{niceyellow}{RGB}{240,228, 66}
\definecolor{niceblue}{RGB}{0,114, 178}
\definecolor{nicevermillion}{RGB}{213,94, 0}
\definecolor{nicereddishpurple}{RGB}{204,121, 167}

\newcommand{\inv}{^{-1}}
\newcommand{\tr}{^{\!\top}}
\newcommand{\invtr}{^{-\!\top}}
\newcommand{\invtrs}{^{-\!\top\slash2}}
\newcommand{\invs}{^{-1\slash2}}
\newcommand{\argmax}{\operatornamewithlimits{argmax}}
\newcommand{\argmin}{\operatornamewithlimits{argmin}}
\newcommand{\diag}{\mathit{diag}}
\newcommand{\se}{\mathrm{se}(3)}
\newcommand{\SE}{\mathrm{SE}(3)}
\newcommand{\SO}{SO(3)}
\newcommand{\R} {{\rm I\!R}}
\newcommand{\E} {{\rm I\!E}}
\newcommand{\bl}{{\bar l}}
\newcommand{\eqdef}{\vcentcolon=}
\newcommand{\algrule}[1][.5pt]{\par\vskip.5\baselineskip\hrule height #1\par\vskip.5\baselineskip}

\newcommand{\rot}[4]{\prescript{#1}{#3}{\mathbf{R}}^{#2}_{#4}}
\newcommand{\tran}[4]{\prescript{#1}{#3}{\mathbf{t}}^{#2}_{#4}}

\newcommand{\campose}[2]{\prescript{#1}{}{\mathbf{X}}_{#2}}
\newcommand{\objpose}[2]{\prescript{#1}{}{\mathbf{L}}_{#2}}
\newcommand{\worldf}{W}
\newcommand{\cammotion}[3]{\prescript{#1}{#2}{\mathbf{T}}_{#3}}
\newcommand{\objmotion}[3]{\prescript{#1}{#2}{\mathbf{H}}_{#3}}
\newcommand{\othmotion}[4]{\prescript{#1}{#2}{\mathbf{#3}}_{#4}}
\newcommand{\objf}{L}
\newcommand{\camf}{X}
\newcommand{\imgf}{I}
\newcommand{\mpoint}[2]{\prescript{#1}{}{\mathbf{m}}_{#2}}
\newcommand{\nhpoint}[2]{\prescript{#1}{}{\tilde{\mathbf{m}}}_{#2}}
\newcommand{\ppoint}[2]{\prescript{#1}{}{\mathbf{p}}_{#2}}
\newcommand{\ipoint}[2]{\prescript{#1}{}{\mathbf{p}}_{#2}}
\newcommand{\icorre}[2]{\prescript{#1}{}{\tilde{\mathbf{p}}}_{#2}}
\newcommand{\opflow}[1]{\prescript{#1}{}{\bm{\phi}}}

\newcommand{\suchthat}{\;\ifnum\currentgrouptype=16 \middle\fi|\;}

\newcommand{\factor}[2]{\lVert {#1} \rVert^2_{\Sigma_{{#2}}}}

\newcommand{\ztwod}{\mathbf{z}_{\text{2D}}}
\newcommand{\zthreed}{\mathbf{z}_{\text{3D}}}

\title{\LARGE \bf Training Trajectory Predictors Without Ground-Truth Data}

\author{Mikolaj Kliniewski\mbox{*}\thanks{\mbox{*} Corresponding author}, Jesse~Morris, Ian~R.~Manchester, and~Viorela~Ila%
\thanks{The authors are with the Australian Centre for Robotics and School of Aerospace, Mechanical and Mechatronic Engineering, University of Sydney, Australia.
	{\tt \{mikolaj.kliniewski, jesse.morris, ian.manchester, viorela.ila\}@sydney.edu.au}}}

\maketitle

\begin{abstract} 
This paper presents a framework capable of accurately and smoothly estimating position, heading, and velocity. Using this high-quality input, we propose a system based on Trajectron++~\cite{salzmann2020trajectron++}, able to consistently generate precise trajectory predictions. Unlike conventional models that require ground-truth data for training, our approach eliminates this dependency. Our analysis demonstrates that poor quality input leads to noisy and unreliable predictions, which can be detrimental to navigation modules. We evaluate both input data quality and model output to illustrate the impact of input noise. Furthermore, we show that our estimation system enables effective training of trajectory prediction models even with limited data, producing robust predictions across different environments. Accurate estimations are crucial for deploying trajectory prediction models in real-world scenarios, and our system ensures meaningful and reliable results across various application contexts.
\end{abstract}


\section{Introduction}

Intelligent vehicles are becoming an integral part of everyday life, making safety a critical concern. One key module that enhances an autonomous vehicle's ability to react in its environment is trajectory prediction: the task of anticipating future poses of dynamic objects. 
By incorporating anticipated movements of dynamic objects into path planning and control algorithms, autonomous systems can navigate more effectively in dynamic environments~\cite{Finean2023ras}. At its core, trajectory prediction models rely on historical positional data of observed objects to predict their future paths. This information forms the necessary part of the input.

The quality of this input data is vital. Current state-of-the-art models are trained and evaluated using estimation data provided by publicly available autonomous driving datasets~\cite{guo2019safe, liu2024survey, yurtsever2020survey}. However, these datasets introduce significant limitations: estimation data is often derived using varying sensor modalities and post-processing methods~\cite{zheng2015trajectory}. This variability leads to inconsistencies across datasets, as similarly formatted data may have subtle but impactful differences in underlying attributes. Consequently, trajectory prediction models trained on one dataset often fail to generalize effectively to another, limiting their robustness in real-world applications.

For real-world deployment, a consistent and accurate estimation module is essential. Dynamic SLAM frameworks~\cite{morris2025dynosam, judd2024ijrr_mvo, gonzalez2023twistslam++} provide a promising solution by converting raw sensor inputs into precise pose and motion estimations, forming the foundation for trajectory forecasting. Unlike the dataset provided ground-truth data that undergoes extensive pre-processing within the dataset as well as often before each prediction model individual training, accurate dynamic SLAM outputs are generated directly from sensor measurements, reducing latency and avoiding the errors introduced during pre-processing.

To address these challenges, we propose a system that integrates a dynamic SLAM module to estimate objects' poses and motions directly from raw sensor data. These estimations are used as input for training, testing, and evaluating trajectory prediction models, eliminating the need for ground truth data during the training process. By relying on data generated by our system, we ensure consistent input quality across different environments, mitigating the variability seen in existing datasets. Moreover, this direct pipeline enhances real-time decision-making by reducing computational overhead caused by additional pre-processing stages.

 This paper introduces a novel framework for direct trajectory prediction of multiple objects utilizing accurate multi-motion estimation from a dynamic SLAM system. To assess the reliability of our approach, we systematically compare the estimation data generated by our system with that from publicly available datasets, evaluating both consistency and accuracy. Additionally, we analyze the impact of our estimation data on prediction performance by comparing trajectory prediction models trained and tested with our data against those using dataset-provided data. Furthermore, we propose a new evaluation metric designed to measure prediction consistency across consecutive time steps, enhancing the robustness of trajectory prediction analysis.

Finally, we argue that trajectory prediction and motion estimation are interdependent modules that should not be studied in isolation. Improvements in one can enhance the other, highlighting the potential benefits of integrating these systems. Our approach serves as a step forward in building robust, real-time, and generalizable trajectory prediction systems for autonomous vehicles.

\section{Related Work}

Dynamic SLAM methods extend traditional SLAM by integrating measurements of dynamic objects, enabling simultaneous estimation of sensor pose, the static background, and object states within the environment.
The state estimation problem for objects can vary depending on the modeling approach.
Some methods choose to model the object state using simple geometric primitives such as cuboids~\cite{gonzalez2023twistslam++}, ellipsoids~\cite{nicholson18cvpr} or quadrics~\cite{Tian2024its_dynaquadric}. 
While these approaches successfully estimate object poses, they do not account for motion or velocity.

In contrast, other methods focus explicitly on estimating object motion~\cite{morris2025dynosam, morris2024icra, judd2024ijrr_mvo, gonzalez2022twistslam}, which is commonly represented as a 3D rigid-body transformation. This approach also allows highly accurate object pose to be recovered. 
Of these of these methods~\cite{morris2025dynosam} reports the best accuracy in terms of object pose and motion estimation. 

Trajectory prediction methods can be broadly categorized into physics-based, learning-based, and hybrid approaches~\cite{huang2022survey, korbmacher2022review, wang2019exploring}. Classical physics-based models, such as constant velocity and constant acceleration for kinematic models, and the bicycle model for dynamic models, provide interpretable and computationally efficient solutions. These methods rely on motion equations to extrapolate future states, making them effective for short-term predictions. However, they often neglect complex agent interactions and environmental data, limiting their applicability in urban traffic scenarios. To address these limitations, modern trajectory prediction methods have increasingly adopted data-driven approaches.
While classic machine learning and reinforcement learning methods remain competitive, they are often outperformed by deep learning approaches~\cite{mozaffari2020deep}.

Combining data-driven methods with physics-based models has emerged as a promising direction, improving prediction accuracy by leveraging spatiotemporal data while maintaining physical realism through constraints and dynamic models. Recent state-of-the-art models reflect this trend. Pedestrian trajectory prediction models often focus on socially-aware predictions~\cite{yue2022human, gupta2018social}, while models for urban environments with both pedestrians and vehicles utilize diverse input sources, such as High-Definition (HD) maps~\cite{salzmann2020trajectron++, lee2017desire, schafer2024caspnet++, feng2024unitraj}.

The challenge of domain shift across datasets and generalization across varying road conditions, regions, and driving styles persists. Ivanovic~\etal\cite{ivanovic2023AdaptivePrediction} address this by introducing an adaptive layer for efficient domain transfer through offline fine-tuning and online adaptation. Feng~\etal\cite{feng2024unitraj} tackle the issue by enhancing dataset size and diversity, leveraging ScenarioNet~\cite{li2024scenarionet} to unify data formats, perform single-step preprocessing, and harmonize discrepancies such as varying recording frequencies, object observation durations, and map annotation precision. Notably, all these state-of-the-art models rely on ground truth data for training.

In this work, we bridge the gap between pose estimation, motion estimation, and trajectory prediction by integrating the entire process from raw sensor data to trajectory forecasting. Unlike existing methods that rely on preprocessed pose data, our unified system ensures consistency across environments without external dependencies. To our knowledge, this is the first approach directly linking sensor inputs, estimation, and prediction, making it well-suited for real-world deployment in diverse settings.

\section{Methodology}

\begin{figure*}[t]
	\centering
    \includegraphics[width=\textwidth]{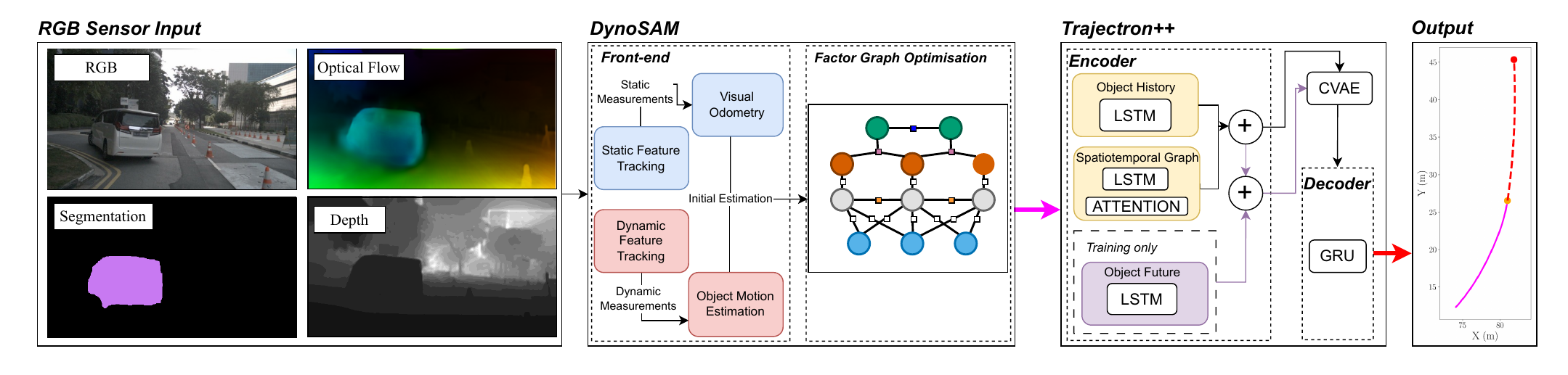}
	\caption{\small{System diagram. Our pipeline processes RGB-D sensor data through the DynoSAM estimation module, which computes object estimates. The estimated position, heading, and velocity are fed into the Trajectron++ model that generates future 2D positions of the object.}}
    \label{fig:system_diagram}
    \vspace{-5mm}
\end{figure*}

\subsection{System architecture}
The system architecture consists of a front-end module that processes raw RGB-D data to track static and dynamic point measurements and estimate vehicle's initial pose as well as the  motion of surrounding objects. These initial estimations are further refined using a Dynamic SLAM back-end, which considerably increases accuracy of the vehicle and object trajectories. The estimated trajectories can serve as inputs to a trajectory prediction model, either during training or at inference time. This architecture (\figref{fig:system_diagram}) guarantees robust and high-accuracy localization, along with motion forecasting of both the vehicle and surrounding dynamic objects.

\subsection{Estimation Module}

DynoSAM~\cite{morris2025dynosam} is a state-of-the-art dynamic SLAM framework developed in our group that uses probabilistic inference to obtain accurate multi-object motion estimation. Built on factor graph principles and implemented with GTSAM~\cite{gtsam}, it efficiently estimates real-time trajectories of both the vehicle and dynamic objects in its environment.

\subsubsection{DynoSAM's front-end}
DynoSAM's front-end (\figref{fig:system_diagram}) is able to distinguish between static and dynamic elements in a scene, estimate initial camera pose and objects' motions, as well as track static and dynamic points. 
The current implementation uses RGB-D sensor data and can be easily extended to LIDAR + RGB data. The separation of static and dynamic objects is achieved by combining instance level semantic segmentation~\cite{JocherUltralyticsYOLO2023} and optical flow~\cite{teed2020raft} with a robust tracking mechanism.

At each time-step, a sparse set of detected static keypoints are tracked across consecutive frames using optical flow.

For dynamic objects, we seek to track a dense set of features for each object. Dense tracking is important to achieve good coverage over the entire observable object for robust motion estimation~\cite{zhang20iros}. The instance segmentation mask is used to associate object labels with pixel measurements. In order to guarantee that the instance labels will be temporally consistent, object masks are tracked using the method in~\cite{zhang2022bytetrack}.

Following the work of~\cite{zhang2020vdoslam}, our front-end uses optical flow to establish initial temporal matching which is further refined using a joint estimation of optical flow and camera pose and object motions. This step is important to ensure robust and accurate feature tracking on both static and dynamic features~\cite{zhang20iros}. 

\subsubsection{DynoSAM's back-end}
The key feature of the estimation module is its focus on estimating the motion of objects rather than their pose. The significance of this approach is analyzed in our previous works \cite{morris2024icra, morris2025dynosam}.

After tracking both static and dynamic features, the initial estimates of camera pose and object motions, along with the static and dynamic maps, are jointly refined using a factor-graph formulation of the inference problem. Depending on the application, the back-end supports various Dynamic SLAM formulations for estimation. Our system enables direct estimation of \textit{world-centric object poses} at each timestep or, alternatively, the \textit{world-centric $SE(3)$ motion}. While directly estimating for the object pose seems a more intuitive choice, the estimation of world-centric $SE(3)$ motion demonstrates superior performance~\cite{morris2025dynosam}. The pose of every object can easily be recovered using:
\begin{equation}
    \objpose{\worldf}{k} = \objmotion{\worldf}{k-1}{k} \: \objpose{\worldf}{k-1}
\end{equation}
where $\objpose{\worldf}{k-1}$ and $\objpose{\worldf}{k}$ is the object pose at timestep $k-1$ and $k$ respectively, and $\objmotion{\worldf}{k-1}{k}$ is the \textit{estimated object motion}.

To ensure seamless integration of DynoSAM for trajectory prediction, we clearly define its output:
\begin{equation}
\mathcal{O}_k = [ \campose{\worldf}{k},   \othmotion{\worldf}{k-1}{\mathcal{H}}{k}, ^{\worldf}\mathcal{L}_{k} , {^\worldf\mathcal{M}_k}]\: k\in \mathcal{K}
\label{equ:dynosam_output} 
\end{equation}
 and is independent of the specific formulation used to solve the estimation problem.  In~\eqref{equ:dynosam_output}, $\campose{\worldf}{k}$ represents the estimated camera pose,   $\othmotion{\worldf}{k-1}{\mathcal{H}}{k}$ and $^{\worldf}\mathcal{L}_{k}$ represent the set of all object motions and poses per-timestep and ${^\worldf\mathcal{M}_k}$ is the set of static and dynamic map points. 
While DynoSAM's back-end is capable of full 3D estimation, existing trajectory-based approaches are constrained to 2D. As a result, in this system variant, we project the estimated motion onto a 2D plane to maintain compatibility.

\subsection{Trajectory Prediction Module}

Our trajectory prediction is based on Trajectron++~\cite{salzmann2020trajectron++}, a graph-structured recurrent neural network designed for multi-agent trajectory prediction in dynamic environments. The model encodes object history and current observations into representation vectors using Long Short-Term Memory (LSTM) networks. Agent interactions are modeled via a directed graph, where nodes represent dynamic objects and edges describe interactions. The graph structure accounts for varying perception ranges (e.g., pedestrians vs. drivers), with $l_{2}$ distance and predefined thresholds determining interaction influence. Edge information from the same semantic class is aggregated and processed by shared-weight LSTMs. The resulting influence representation vector, enhanced by the attention mechanism, is concatenated with the node history vector to form a unified node representation. Additional data, such as HD maps or LIDAR, can be incorporated by encoding and concatenating them with these vectors.

Trajectron++ employs a Conditional Variational Autoencoder (CVAE) framework. During training, a bi-directional LSTM encodes the agent's future ground-truth trajectory, producing a latent variable. This, along with the representation vector, is fed into Gated Recurrent Unit (GRU) decoder. The GRU outputs parameters for a bivariate Gaussian distribution over control actions (e.g., acceleration, steering rate), which are integrated with agent dynamics to generate position-space trajectories. This structure enables accurate predictions in complex, dynamic scenes. For further details, see~\cite{salzmann2020trajectron++} and~\cite{ivanovic2023AdaptivePrediction}.

Trajectron++ was chosen for its strong benchmark performance, real-time capabilities, and adaptability to diverse data types. At its core, it uses 2D state information (position, velocity, orientation) and agent category to respect motion constraints. Pedestrians are modeled as single integrators, and vehicles as dynamically-extended unicycles~\cite{lavalle2006better}.

For our system, we trained the model from scratch using the adaptive prediction framework~\cite{ivanovic2023AdaptivePrediction} with default hyperparameters, opting for unimodal predictions. Given the limited training data and to maintain its characteristics, we avoided noise augmentation. We used a sampling rate of \SI{20}{\hertz}, with input history trajectories spanning between $1$ to $6$ frames and predicted trajectories extending over $30$ frames, corresponding to a \SI{1.5}{\second} prediction horizon. Physical constraints were applied to ensure realistic motion predictions, limiting vehicle turn rate to $±0.7$\,rad/s and the acceleration to $±4$\,$\text{m/s}^{2}$.

Unlike prior approaches that rely on ground-truth or pre-processed data, we trained and tested our model directly on raw data obtained from our estimation module. This ensures that the inputs are consistent across all datasets, as they are generated by the same system, enhancing robustness and applicability in real-world scenarios. By eliminating the need for artificially curated training data, our method better captures the complexities and uncertainties inherent in real-world motion estimation.

\section{Experiments}

The goal of our experiments is to validate that our system can effectively facilitate both training and inference using multi-object motion estimation derived from DynoSAM, even surpassing models trained on ground truth data. For that we train and test our variant of Trajectron++ model using the output from DynoSAM, using the ground-truth (GT) provided by the data set and ground-truth values processed using an Extended Kalman Filter (GT+EKF) to examine the effect of noise smoothing.

The trajectory predictions at each step are compared with the actual trajectories generated by the same method used as input (e.g. estimated by DynoSAM, GT or GT+EKF). These comparisons allow us to evaluate the usability, smoothness, and impact of DynoSAM on trajectory prediction accuracy.

\begin{table}[t]
\footnotesize
\centering
\setlength{\tabcolsep}{2.8pt}
\caption{\small{Description of eligible data from selected datasets.}}
\label{tab:data_source}
\begin{tabular}{c|ccccccccc|c|c}
\toprule
 & \multicolumn{10}{c|}{KITTI} & \multicolumn{1}{c}{nuScenes}\\
 & $00$ & $01$ & $02$ & $03$ & $04$ & $05$ & $06$ & $18$ & $20$ & sum & $61$ \\
\midrule
\midrule
 eligible objects & $2$ & - & - & $2$ & - & $1$ & - & $2$ & $8$ & $15$ & $1$ \\
 \midrule
 testing instances & $144$ & - & - & $67$ & - & $81$ & - & $47$ & $183$ & $522$ & $149$ \\
\bottomrule
\end{tabular}
\vspace{-4mm}
\end{table}

\subsection{Dataset}
The KITTI dataset is widely used for testing multi-object tracking and dynamic SLAM systems, making it an ideal choice for our experiments. It provides the necessary RGB-D sensor data for our estimation system and the center of the bounding box for each visible object, which we use as the ground-truth pose for trajectory prediction based on GT data.
Not all objects can be evaluated. An object in the dataset becomes eligible for our study if it appears in at least $32$ frames, calculated by including a minimum of 1 frame of historical data, the current frame, and a $30$-frame prediction horizon. The dataset encompasses diverse driving scenarios, including various road types, traffic conditions, and environments.
Recorded in a real-world dynamic environment at \SI{20}{\hertz}, the dataset consists of $9$ sequences, $5$ of which contain observations of eligible objects. As shown in \tabref{tab:data_source}, there are $15$ eligible objects in the KITTI dataset, yielding $522$ testing instances and $1264$ training instances, enhanced through input augmentation.
With its detailed annotations and sensor data, the KITTI dataset is well-suited for evaluating the integration of pose and motion estimation with trajectory prediction in realistic conditions.

To validate our approach, we incorporate sequence $61$ from the nuScenes mini dataset, which contains an object exhibiting a trajectory analogous to those observed in KITTI sequence $00$. For scene depth estimation needed for DynoSAM, we leverage the state-of-the-art Dynamo-Depth model~\cite{sun2024dynamo}. Since our estimation pipeline operates without prior scale knowledge, we align the data by applying a uniform scaling factor to both positional coordinates and velocity estimates, calibrating them to match the velocity profile of Object $2$ in KITTI $00$. Although the data stream frequency fluctuates between \SI{10}{\hertz} and \SI{20}{\hertz} during estimation, due to hardware-triggered camera operation in nuScenes, we assume a fixed rate of \SI{20}{\hertz} to ensure compatibility with models trained on the KITTI dataset.

\subsection{Metrics}
Our evaluation focuses on two aspects: training and inference \textit{input data quality} and \textit{prediction performance}.

\subsubsection{Estimation Data Evaluation}
We asses the accuracy of DynoSAM by evaluating camera and object motion estimation in $\SE$ using Relative Pose Error ($\mathbf{RPE}$) as defined by Sturm~\etal\cite{sturm2012iros}.
The results are computed and reported using the root-mean-squared error ($\mathbf{RMSE}$) of the translation ($\mathbf{RPE}_t$) and rotation ($\mathbf{RPE}_r$) components separately.

\subsubsection{Prediction Evaluation}

To evaluate the accuracy of the predictions, we use the standard metrics of Average Displacement Error ($\mathbf{ADE}$) and Final Displacement Error ($\mathbf{FDE}$). These are the most commonly used metrics in trajectory prediction tasks~\cite{huang2022survey, mozaffari2020deep} and provide insight into both overall and endpoint accuracy. The $\mathbf{ADE}$ is an $l_{2}$ distance between the predicted trajectory and the ground truth:
\begin{equation}
    \mathbf{ADE} = \frac{1}{T} \sum_{t=1}^{T} | Y^{t}_{\text{pred}} - Y^{t}_{\text{GT}} |  \text{,} 
\end{equation}
where $T$ represents the number of steps in the prediction horizon, and $Y^{t}_{\text{pred}}$ and $Y^{t}_{\text{GT}}$ are predicted and corresponding ground truth locations $t$ steps in the future.
Similarly, the $\mathbf{FDE}$ is the $l_{2}$ distance between the final predicted position and the corresponding ground truth location:
\begin{equation}
    \mathbf{FDE} = | Y^{T}_{\text{pred}} - Y^{T}_{\text{GT}} |  \text{,} 
\end{equation}
where $T$ denotes a final predicted time step.

Additionally, we introduce a novel evaluation metric to assess model confidence, the Absolute Consistency Error ($\mathbf{ACE}$). 
$\mathbf{ACE}$ is the $l_{2}$ distance between the last and second to last steps in the prediction horizon, that is $T$ and $T-1$, made at two consecutive time steps:
\begin{equation}
    \mathbf{ACE} = | Y^{T}_{\text{pred} \space (s)} - Y^{T-1}_{\text{pred} \space (s+1)} |  \text{,}
\end{equation}
where $s$ is the time when predictions are made, $Y^{T}_{\text{pred} \space (s)}$ denotes a prediction made at time $s$ forecasting position at $T$ steps forward, and $Y^{T-1}_{\text{pred} \space (s+1)}$ denotes a prediction made at time $s+1$ forecasting position at $T-1$ steps forward.
The $\mathbf{ACE}$ metric quantifies temporal consistency, revealing deviations that may indicate uncertainty or instability. Unlike $\mathbf{ADE}$ and $\mathbf{FDE}$, it captures sequential prediction variations, providing a more reliable measure of prediction usability in planning and navigation. 

\subsection{Results}

\subsubsection{Estimation Component}

\begin{table}[t]
\footnotesize
\centering
\setlength{\tabcolsep}{2.0pt}
\caption{\small{Quantitative evaluation of 3D camera trajectory and object poses generated by DynoSAM.}}
\label{tab:estimation_results}
\begin{tabular}{c|ccccccccc|c}
\toprule
 & \multicolumn{10}{c}{KITTI} \\
 & $00$ & $01$ & $02$ & $03$ & $04$ & $05$ & $06$ & $18$ & $20$ & avg.  \\
\midrule
\midrule
camera $\text{RPE}_r$(\si{\degree}) & $0.04$ & $0.03$ & $0.02$ & $0.05$ & $0.06$ & $0.05$ & $0.05$ & $0.04$ & $0.04$ & $0.04$  \\
\midrule
camera $\text{RPE}_t$(\si{\meter}) & $0.04$ & $0.04$ & $0.03$ & $0.05$ & $0.06$ & $0.05$ & $0.01$ & $0.04$ & $0.02$ & $0.04$\\
\midrule
\midrule
object $\text{RPE}_r$(\si{\degree}) & $1.38$ & $0.80$ & $1.06$ & $0.27$ & $1.04$ & $0.62$ & $2.74$ & $1.16$ & $0.33$ & $1.04$  \\
\midrule
object $\text{RPE}_t$(\si{\meter}) & $0.27$ & $0.32$ & $0.79$ & $0.19$ & $0.92$ & $0.16$ & $0.48$ & $0.2$ & $0.12$ & $0.38$\\
\bottomrule
\end{tabular}
\vspace{-2mm}
\end{table}

The \tabref{tab:estimation_results} represents the evaluation of the DynoSAM estimation, particularly the camera and objects motion estimation assessed using rotational and translational components of $\mathbf{RPE}$ for both the estimated camera and objects trajectories. We present these findings here for completeness. 
For detailed evaluation of the estimation system and comparison against other state of the art estimation methods, we encourage readers to get familiar with DynoSAM~\cite{morris2025dynosam}.

\begin{figure}[!htb]
    \centering
  \includegraphics[width=0.9\columnwidth]{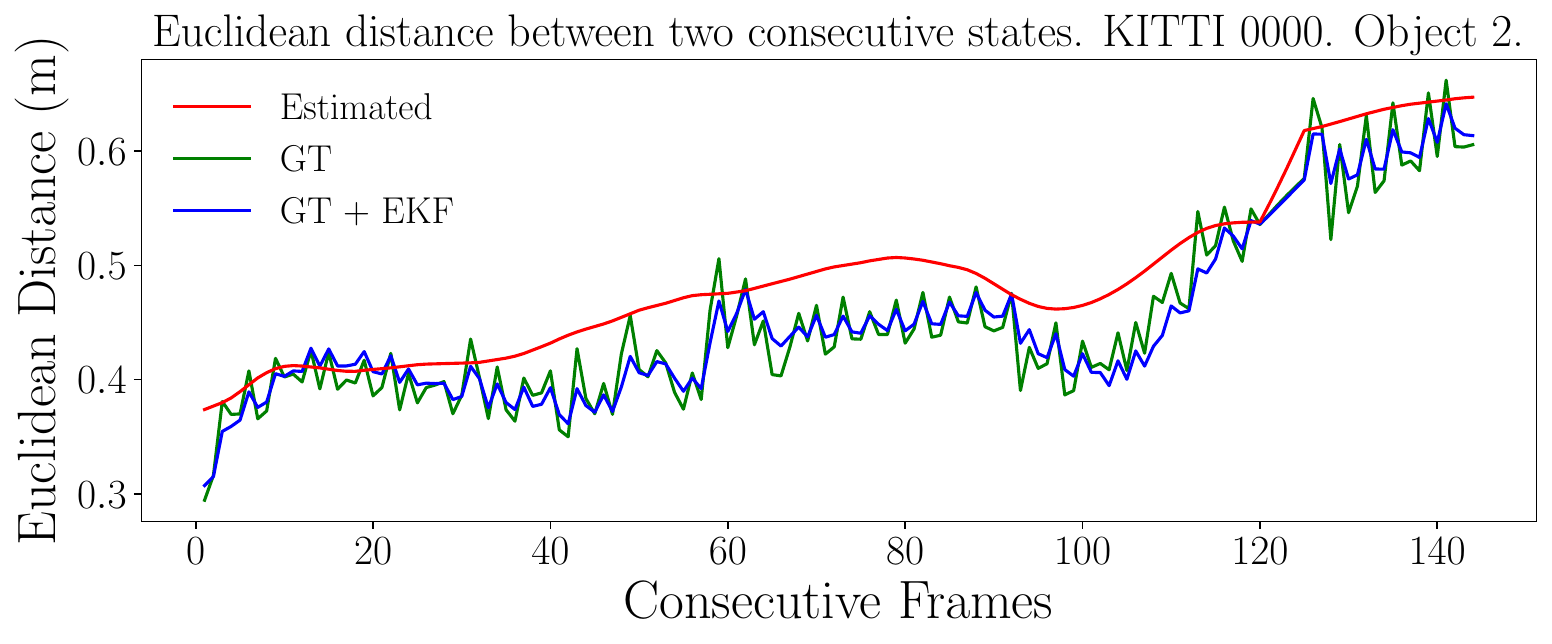}
   \caption{\small{Euclidean distance between consecutive object positions across the trajectory for three data sources. Noisier curves indicate less physically feasible estimations.}}
    \label{fig:xy}

    \centering
  \includegraphics[width=0.9\columnwidth]{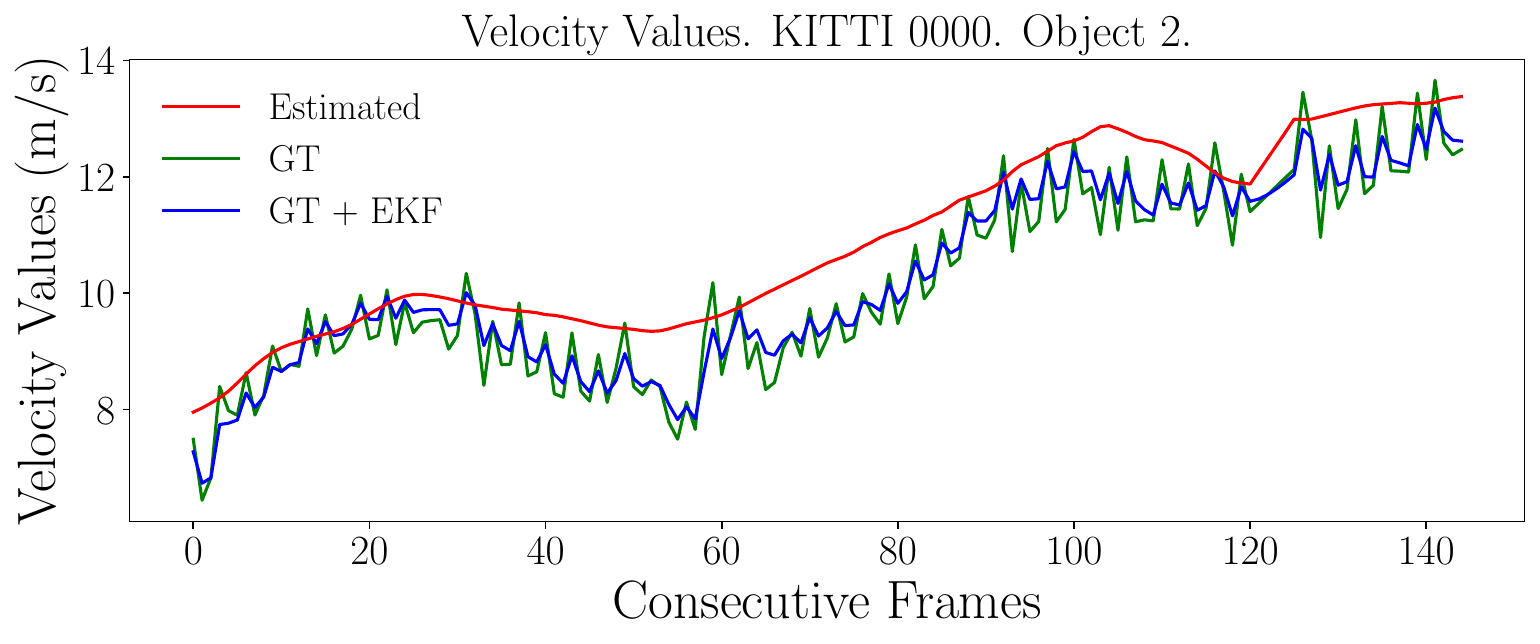}
   \caption{\small{Object's velocity values across the trajectory for three data sources. Noisier curves indicate less physically feasible estimations.}}
    \label{fig:vel}
    
    \centering
  \includegraphics[width=0.9\columnwidth]{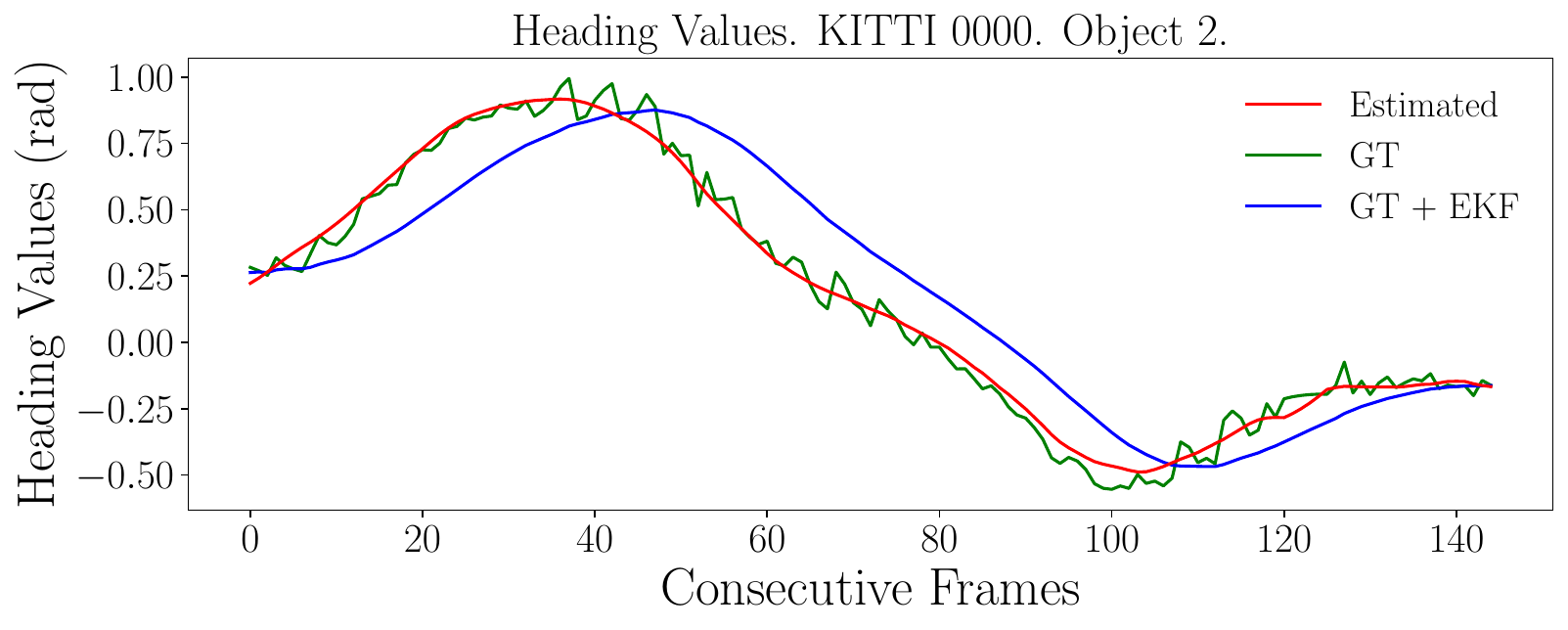}
   \caption{\small{Object's heading values across the trajectory for three data sources. Noisier curves indicate less physically feasible estimations.}}
    \label{fig:heading}
\end{figure}

We assess the smoothness of data sources by analyzing Euclidean distances between consecutive 2D positions and changes in heading and velocity. This allows us to compare estimation data with ground truth, clarifying inputs for trajectory prediction. To illustrate, we visualize object $2$ from KITTI sequence $00$, as it appears in the most consecutive frames, ensuring clearer interpretation.
\figref{fig:xy} shows the Euclidean distance between consecutive positions, while \figref{fig:vel} and \figref{fig:heading} depict velocity and heading values over time. \figref{fig:xy} and \figref{fig:vel} help assess movement naturalness, as real vehicles accelerate and decelerate gradually. Large, abrupt changes, especially at a high frame rate, indicate poor estimation. Similarly, \figref{fig:heading} highlights unrealistic fluctuations in direction.
Results confirm DynoSAM’s inherent smoothness. The ground truth raw data exhibits unnatural oscillations across all metrics, while the smoothed ground truth improves heading continuity but still contains noise in velocity and position metrics.

\subsubsection{Trajectory Prediction}

\begin{table}[t]
\footnotesize
\centering
\setlength{\tabcolsep}{3.5pt}
\caption{\small{Quantitative evaluation of trajectory predictions for $T_{pred}= 30 \ \text{steps} =  1.5s$ obtained using different data sources. Models trained using data from our estimation method outperform ground truth and smoothed ground truth data models (lower is better).}}
\label{tab:results_prediction}
\begin{tabular}{c|c|ccccc|c|c}
\toprule
 & & \multicolumn{6}{c|}{KITTI}& \multicolumn{1}{c}{nuScenes}\\
 & & $00$ & $03$ & $05$ & $18$ & $20$ & avg. & $61$\\
\midrule
\midrule
\multirow{3}{*}{ADE} & Ours & $\mathbf{1.15}$ & $\mathbf{0.47}$ & $\mathbf{0.99}$ & $0.44$ & $\mathbf{0.55}$ & $\mathbf{0.76}$ & $0.92$\\
 & GT & $1.61$ & $0.99$ & $1.44$ & $0.49$ & $0.77$ & $1.11$ & - \\
 & GT + EKF & $1.45$ & $0.86$ & $1.4$ & $\mathbf{0.39}$ & $0.69$ & $1.00$ & - \\
\midrule
\multirow{3}{*}{FDE} & Ours & $\mathbf{3.18}$ & $\mathbf{1.36}$ & $\mathbf{2.74}$ & $1.27$ & $\mathbf{1.6}$ & $\mathbf{2.15}$ & $2.58$\\
 & GT & $4.18$ & $2.36$ & $3.83$ & $1.18$ & $2.12$ & $2.90$ & - \\
 & GT + EKF & $3.89$ & $2.35$ & $3.76$ & $\mathbf{0.93}$ & $1.98$ & $2.73$ & - \\
\midrule
\multirow{3}{*}{ACE} & Ours & $\mathbf{0.39}$ & $\mathbf{0.26}$ & $\mathbf{0.21}$ & $0.71$ & $0.8$ & $\mathbf{0.51}$ & $0.3$ \\
 & GT & $0.76$ & $1.48$ & $0.56$ & $0.65$ & $0.96$ & $0.88$ & - \\
 & GT + EKF & $0.67$ & $1.27$ & $0.46$ & $\mathbf{0.39}$ & $\mathbf{0.79}$ & $0.73$ & - \\
\bottomrule
\end{tabular}
\vspace{-2mm}
\end{table}

\begin{figure}[t]
    \centering
  \includegraphics[width=0.9\columnwidth]{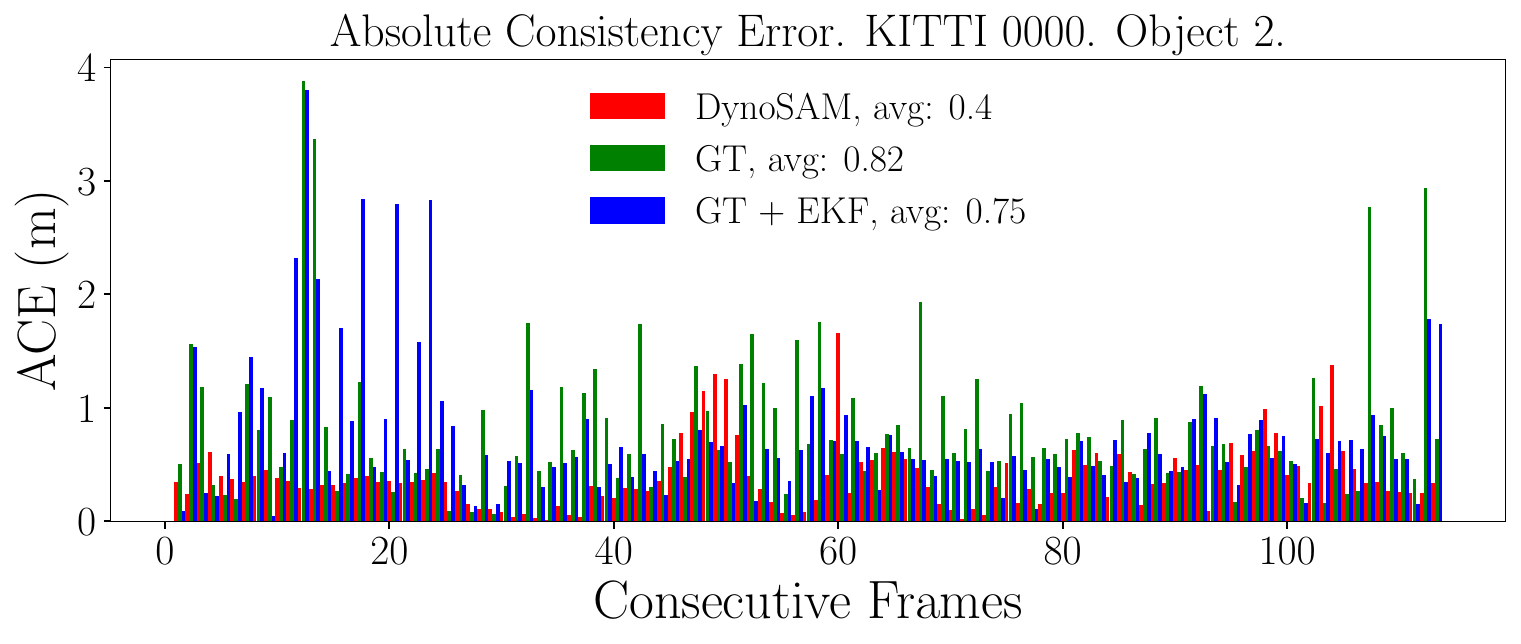}
   \caption{\small{$\mathbf{ACE}$ of object $2$ in KITTI $00$ per three models.}}
    \label{fig:ace}
    \vspace{-3mm}
\end{figure}

\begin{figure}
    \centering
  \includegraphics[width=0.9\columnwidth]{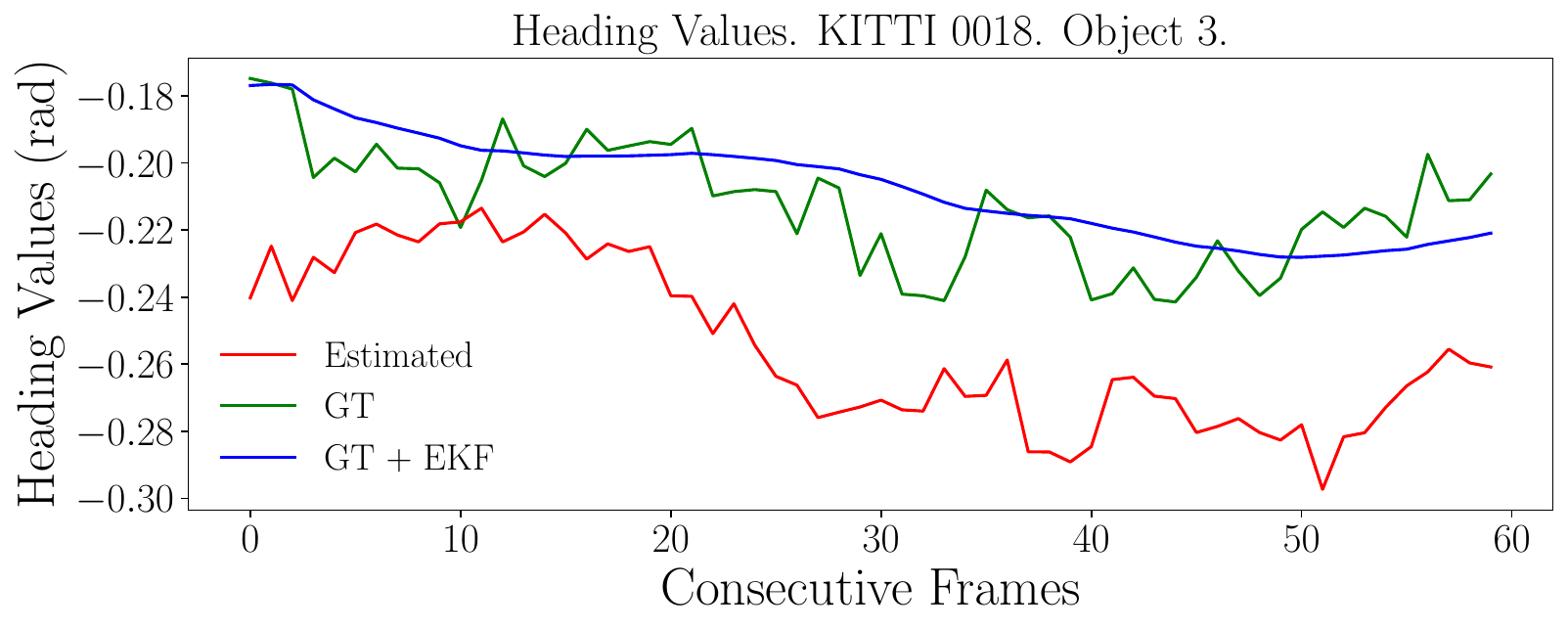}
   \caption{\small{Object's heading values across the trajectory for three data sources. Noisier curves indicate less physically feasible estimations.}}
    \label{fig:heading18}
\end{figure}

Due to the limited number of objects observed for at least $32$ consecutive frames, as reported in \tabref{tab:data_source}, we evaluate the predictions using one nuScenes mini, and five KITTI sequences. For each \textit{Estimated}, \textit{GT}, and \textit{GT+EKF} case, we trained the model using data from all other available KITTI sequences before performing inference on a target sequence. For the nuScenes sequence evaluation, the model was trained on all estimated KITTI data.
We compare our predictions to the future trajectory of an object, which is derived using the same method employed to obtain the data for both training and prediction.
The results are available in \tabref{tab:results_prediction}. The `avg.' column in the table describes the weighted average of the error; that is the average $\mathbf{ADE}$, $\mathbf{FDE}$ and $\mathbf{ACE}$ errors per trajectory prediction in KITTI. The $\mathbf{ADE}$ and $\mathbf{FDE}$ metrics were all together measured $522$ times, whereas $\mathbf{ACE}$ metric was measured $507$ times as two consecutive predictions are needed for the error calculation. 
\figref{fig:ace} represents the $\mathbf{ACE}$ of predictions of object $2$ in sequence $00$. Model trained and tested using \textit{Estimated} data significantly outperforms the models trained and tested using \textit{GT} and \textit{GT+EKF}, respectively, in all sequences except KITTI $18$. Sequence $18$ contains 2 objects moving in nearly straight paths. As shown in \figref{fig:heading18}, both the \textit{Estimated} and \textit{GT} data exhibit a minor noise in the heading values while \textit{GT+EKF} smooths these values, yielding better results. This highlights the importance of input data smoothness in enhancing trajectory prediction accuracy.

\section{Conclusion and Future Work}

We proposed the first system for estimating accurate data for both training and inference of trajectory prediction models. By leveraging DynoSAM, our approach generates smooth, high-quality data from diverse datasets, enabling accurate predictions even with limited training data. Unlike traditional methods that rely on artificially curated datasets, our system learns directly from realistic, potentially noisy inputs, enhancing its robustness and applicability in real-world environments. This makes it particularly valuable for path planning and control systems, where anticipating object motion is critical for collision avoidance and efficient navigation. Furthermore, our system’s consistent predictions ensure stability, allowing navigation algorithms to adapt smoothly without abrupt changes. Overall, our work represents a significant step toward end-to-end trajectory prediction directly from raw sensor data, paving the way for safer and more reliable autonomous systems.

While Trajectron++ currently operates on 2D data, we are exploring the development of a predictor that can leverage full SE(3) data from DynoSAM, enabling more comprehensive motion estimation. Additionally, Trajectron++ supports map integration, and in our recent work, we use DynoSAM to construct a dynamically updated 3D map of the dynamic scene. Incorporating this evolving spatial representation is expected to significantly enhance prediction accuracy. Moreover, training on larger datasets and testing across diverse sensor configurations will further improve robustness and adaptability. Finally, integrating adaptive prediction models could refine trajectory forecasts across a wider range of scenarios.

\bibliographystyle{IEEEtran}
\bibliography{./IEEEabrv, ./bibliography}

\end{document}